%% file: 3286180_ZhaoStamm.tex
\documentclass[letterpaper,twocolumn,fleqn]{article} 

\usepackage{ist}
\usepackage{cite}
\usepackage{comment}
\usepackage[table]{xcolor}
\usepackage{amsmath,graphicx, amssymb}
\renewcommand{\baselinestretch}{1.0}

\DeclareMathOperator{\E}{\mathbb{E}}

\newcolumntype{K}[1]{>{\centering\arraybackslash}p{#1}}

\title{The Effect of Class Definitions on the Transferability of Adversarial Attacks Against Forensic
CNNs}

\author{ Xinwei Zhao and Matthew C. Stamm; Drexel University; Philadelphia, PA, xz355@drexel.edu, mstamm@coe.drexel.edu}

\date{} 

\begin{document} 

\maketitle 

\thispagestyle{empty} 


\begin{abstract}
\input{sections/abstract.tex}
\end{abstract}

\section{Introduction}
\label{sec:intro}
\input{sections/intro.tex}
\section{Background}
\label{sec:background}
\input{sections/background.tex}
\section{Investigation procedure}
\input{sections/proposed.tex}

\section{Experiments}
\input{sections/experiment.tex}

\section{Conclusion}
\input{sections/conclusion.tex}
\section{Acknowledgment}
\input{sections/acknowledgement.tex}

\vspace{-0.5em}
\small
\bibliographystyle{bibstyle}
\bibliography{refs/citations}

\end{document}

%% file: sections/abstract.tex
In recent years, convolutional neural networks (CNNs) have been widely used by researchers to
perform forensic tasks such as image tampering detection. At the same time, adversarial attacks
have been developed that are capable of fooling CNN-based classifiers. Understanding the
transferability of adversarial attacks, i.e. an attack’s ability to attack a different CNN than the one it
was trained against, has important implications for designing CNNs that are resistant to attacks.
While attacks on object recognition CNNs are believed to be transferrable, recent work by Barni et
al. has shown that attacks on forensic CNNs have difficulty transferring to other CNN architectures
or CNNs trained using different datasets. In this paper, we demonstrate that adversarial attacks on
forensic CNNs are even less transferrable than previously thought – even between virtually identical
CNN architectures! We show that several common adversarial attacks against CNNs trained to
identify image manipulation fail to transfer to CNNs whose only difference is in the class definitions
(i.e. the same CNN architectures trained using the same data). We note that all formulations of
class definitions contain the “unaltered” class. This has important implications for the future design
of forensic CNNs that are robust to adversarial and anti-forensic attacks.

%% file: sections/intro.tex
The integrity and authenticity of multimedia contents are top concerns in many scenarios, such as criminal investigation and news reporting\cite{stamm2013_overview}. Research has shown that many editing operations, such as resizing~\cite{kirchner2008fast} or contrast enhancement~\cite{stamm2010forensic}, will leave unique traces behind. Many forensic algorithms have been developed to detect or identify editing operations~\cite{amerini, fridrich2003detection,bayram2009efficient, siwei, mayer2018accurate,  MISLNet, bianchi2012image, chen2007imaging, splicebuster, JPEG_ghosts, 7169770, 7542518}. In recent years, convolutional neural networks (CNNs) have been widely used by researchers to perform forensic tasks such as image tampering detection\cite{mayer2020similarity, bondi2017cluster, CNN_antiJPEG, MISLNet} and source identification\cite{ tuama2016cameraWIFS, noiseprint, bondi2017_1st_step}.

In some scenarios, an intelligent attacker may attempt to launch an adversarial attacks to fool forensic algorithms~\cite{sharma2016wifsanti, fontani2012hiding_MF, kirchner2008_hiding_resampling, antiContrast2010, stamm_dithering}.  Many adversarial attacks have been found to be able to fool deep learning based algorithms~\cite{DeepFool_2016_CVPR, GAN2014, Papernot_2017, 2014arXiv1412.6572G, CW_attack, evasionattack, biggio2012poisoning, Mislgan, huang2017adversarial}. Researchers have already demonstrated that fast gradient sign method (FGSM)~\cite{counter2017CNN} and generative adversarial network (GAN)~\cite{chen2019generative, Kim2018medianFiltGAN} based attacks can be used to fool forensic CNNs. Therefore, it is important to understand the capability and limitations of the adversarial attacks.

Transferability is one of the well-known problems pertaining to adversarial attacks~\cite{liu2016delving,  papernot2016transferability, papernot2016limitations, barniadversarial}. Transferability issues occur when the attacker attempts to attack a different CNN than the one that were explicitly trained against.  Since many attacks operate by pushing the adversarial examples across the boundaries of the target class, it is important for the attacks to be able to observe the gradient of the target classifier with respect to the input data.  However, when the CNN used to train the attack cannot fully mimic the boundaries of the target CNN, the obtained adversarial examples may not be able to transfer.  Two common reasons that can cause attacks' transferability issues are training data discrepancy and CNN architecture discrepancy. 

Understanding the transferability of adversarial attacks has important security implications. If information can be discovered that negatively effects an attacks transferability, it can be used to defend CNNs against attack. Additionally, knowledge of attack transferability helps researchers understand how feasible real-world adversarial attacks could be.
While previous research has shown that attacks against object recognition CNNs can transfer to
attack CNNs with different architectures or trained using different data, recent research in
multimedia forensics shows an opposite result. Specifically, work by Barni et al. has shown that
attacks on forensic CNNs have difficulty transferring to attack other CNN architectures or CNNs
trained using different datasets~\cite{transferability_Barni}.

In this paper, we demonstrate that adversarial attacks on forensic CNNs are even less
transferrable than previously thought – even between virtually identical CNN architectures!  Particularly,
we discover that several common adversarial attacks against forensic CNNs fail to transfer
between CNNs whose only difference is in the class definitions (i.e. the same CNN architectures
trained using the same data). We note that all formulations of class definitions contain the
“unaltered” class.  To investigate the impact of class definitions on forensic CNNs, we assume that attacker knows every details of the forensic CNNs, including the training data and CNN architecture. The only missing information of the attacker is the class definition. Next,  we defined three typical class definitions for image manipulation forensic CNNs by grouping individual manipulation or parameterization of individual manipulation. Then we use the attacked images that are produced by fooling one forensic CNN to fool the other CNNs whose only difference is the class definition. We defined the successful attack rates (SARs) and transferability scores (T-scores) to measure the success and transferability of adversarial  attacks. By conducting an extensive amount of experiments, we found that adversarial attacks are difficult to transfer to other class definitions of the same CNN architecture. Moreover, a secondary finding of ours is that binary classification of forensic CNNs (i.e grouping all manipulation into one class) performs slightly more robust than the other two class definitions. 
This has important implications for the future design forensic CNNs that are
robust to adversarial and anti-forensic attacks.

%% file: sections/background.tex
We assume that an attacker applies some editing operations to an images and then launches an adversarial attack attempted to bypass the detection. The investigator will  use a forensic  CNN to identify if the image presented was unaltered or not. 

For a single forensic manipulation identification CNN, there exists different ways to form class definitions. For instance, an 
binary decisions of unaltered or manipulated,  multi-class definitions of unaltered vs several individual manipulations, 
 or multi-class definitions of unaltered vs. several parameterized versions of individual manipulations.  Each of the above class definitions includes the “unaltered” class.
\vspace{-1em}
\subsection{Near-perfect knowledge scenario}
Previous research has shown the attacker's knowledge pertaining to the target investigator's algorithm determines how easy and successful attacks can be~\cite{chen2019generative, barniadversarial}.  Therefore, depending on the amount of knowledge accessible to attackers, it is common to  categorize the scenarios into the perfect knowledge scenario and partial knowledge scenarios.  The perfect knowledge scenario is when attackers can observe the  every detail of the investigator's algorithm or they  can obtain an identical copy of the investigator's algorithm. Under the perfect knowledge scenario, attackers can directly integrate the investigator's CNN into their attack and train the attack explicitly bypass the detection of the identification CNN.  All other scenarios are categorized as partial knowledge scenarios. Under partial knowledge scenarios, attackers has no full access to the investigator's CNN. As a result, attackers have to ensure their trained attack is capable of fooling different CNNs than the CNN explicitly trained against. If an attack fails to fool different CNNs, transferability of the attack occurs.  Two common reasons that cause that attack's transferability  are the dependencies of training data and CNN architectures~\cite{transferability_Barni, liu2016delving}.  

To investigate the transferability of adversarial attacks induced by class definition, we formulate a special partial knowledge scenario, the near-perfect knowledge scenario. Under this scenario, the attacker knows every details of the investigator's CNN architecture and also will use identical training data as the investigator. The only missing information of the attacker is the class definition of the target CNN (i.e the attacker does not know how the investigator forms the output classes for the forensic identification CNN.).

%% file: sections/proposed.tex
To investigate the impact of class definition on transferability of adversarial attacker, we used the following procedure:
1)	We categorized three different class definitions that could be used by forensic CNNs attempting to identify image editing.
2)	We trained six different forensic CNNs to perform editing detection and achieve their baseline performance under each class definition.
3)	We implemented two popular adversarial attacks and obtain their Successful Attack Rate (SAR) in the perfect knowledge scenario (without attempting transfer).
4)	We evaluated each attack’s ability to transfer to an identical CNN whose only difference is the class definition used in the near perfect knowledge scenario, then interpreted the results.
A detailed description of our experimental procedure, as well as the metrics used to evaluate the attacks is provided below.

\subsection{Class definitions}
 There are several ways to define the classes used by a forensic CNN created to identify image manipulation.  While all class definitions include the “unaltered class” other classes may differ depending on if different manipulations, as well as different parameterizations of manipulations, are grouped together into one class.  In this work, we consider the following three different CNN class definitions.
 
\underline{\smash{Manipulation detection:}} In this class definition, only two classes are used: “manipulated” and “unaltered”. Any type of editing is grouped together into the “manipulated” class.  This class definition would be used if the investigator only wants to know if an image has been modified in any means. 

\underline{\smash{Manipulation classification:}} In this multi-class case, one class is assigned to “unaltered” along with one class for each individual editing operation.  All parameterizations of that editing operation are grouped together into a single class.  This class definition would be used if the investigator not only wants to know if the image has been modified, but also wants to know the individual manipulation applied to the image. 

\underline{\smash{Manipulation parameterization:}} In this multi-class case, one class is assigned to “unaltered” and separate classes are assigned to each pair of manipulation and parameterization (or range of parameterizations).  For example, median filtering with a 3x3 window would be a separate class than median filtering with a 5x5 window.  This class definition could be used if the investigator wants to know very detailed information about a possible forger or identify inconsistencies in editing within an image.

\subsection{Image forensic CNNs}
In this paper, we examined six well-known CNN architectures, including  MISLnet~\cite{MISLNet}, TransferNet\cite{TransferNet}, PHNet~\cite{PHNet}, SRNet~\cite{SRNet}, DenseNet~\cite{DenseNet} and VGG-19~\cite{VGGnet}. While some of the CNN architectures were initially used for computer vision or steganalysis tasks, they can be adapted to train for image forensics.  

For each CNN architecture, we trained forensic CNNs using the above three class definitions. All CNNs were trained using the same dataset created from the Dresden Image Database (more detail is provided in the results section).  Furthermore, CNNs with the same architecture were trained using the same hyperparameters for all class definitions.

\subsection{Adversarial attacks}
To fool a forensic CNN, images modified by an attack should be classified as “unaltered” by that (or other) CNNs.  As a result, attacks used our work operate in a targeted fashion, where the “unaltered” class is always the attack’s target.

We used two well-known adversarial attacks in our experiments: the iterative targeted fast gradient sign method (I-FGSM) attack and the generative adversarial network (GAN) based attack.  These two attack methods are very commonly used in anti-forensics (as well as the broader ML community), and are described below.

\underline{\smash{Iterative targeted fast gradient sign method (targeted I-FGSM):}} It operates by iteratively adding a small noise to the original image $I$ and to push the adversarial examples $I_{adv}$  to the target classes (i.e unaltered class in this context).  At each iteration, the gradient is calculated with respect to the attacked image produced from previous iteration.  The equation of targeted I-FGSM attacks is,
\vspace{-1em}
\begin{align}
I_{adv}^0 &= I\\
I_{adv}^{n+1} &= I_{adv}^n-\epsilon\times sign\nabla_{I_{adv}^n }J (I_{adv}^n, y_{unaltered})
\end{align}
where $n$ denotes the index of iteration, $\epsilon$ denotes a small number, $J(\cdot)$ denotes the loss function, and $y_unaltered$ denotes target class label. 

\underline{\smash{Generative Adversarial Network (GAN)-Based Attack: }} GAN-based method operates by training a GAN network to obtain a generator and then uses the generator to produce an image that can mimic the statistics of unaltered images. 

A traditional GAN~\cite{GAN2014} is trained using a min-max function, 
\begin{equation}
\min_{G}\max_{D} \E_{I\sim p_r(I)}[\log D(I)]+\E_{I_{adv}\sim p_g(I_{adv})}[\log(1- D(I_{adv}))]
\end{equation}
where $G$ denotes the generator, $D$ denotes the discriminator, $ p_r(I)$ denotes the distribution of unaltered images, 
$p_g(I_{adv})$ denotes the distribution of adversarial images and $\E$ denotes the operation of taking expected value. 

We adopted MISLGAN method which was has been initially designed for fooling camera model identification CNNs\cite{Mislgan}.   MISLGAN is consisted of three major components, a generator, a discriminator and a pre-trained forensic CNN. While the generator and the discriminator are trained in the same fashion as the traditional GAN, the pre-trained is introduced to force the generator to produce an image that can mimic the forensic information of the ``unaltered" image.  To attack the manipulation detection CNNs, we modified MISLGAN by removing the synthetic CFA module in the generator. Due to the page limitation of the paper, we advise the readers to find details about the architecture and loss formulation of MISLGAN in the original paper. 
\vspace{-1em}
\subsection{Evaluation metrics}
We define the successful attack rate and transferability score to evaluate the performance and transferability of the attack against the classifiers.
 
\underline{\smash{Successful attack rate (SAR):}} To evaluate the performance of the anti-forensic crafted images against manipulation detection CNNs, we calculated the percentage that the adversarial images are classified as “unaltered” by each CNN, and we define this percentage as successful attack rate (SAR).  CNNs that have a stronger resistance to an anti-forensic attack should have lower SARs. 

\underline{\smash{Transferability score (T-Score): }} To evaluate an attack’s transferability, we calculated transferability score as the SAR of the unknown classifier over the SAR of the known classifier. The known classifier is directly used when launching the attack and the unknown classifier is used for classifying the adversarial images created by the attack.  As a result, when an attack has good transferability, the transferability score should be high. Otherwise, the transferability would be low. For example, when all adversarial images produced by fooling one forensic CNN can fool other unseen CNNs,  the transferability score equals to 1. We would like to point out that the transferability score should be positive and can be higher than 1. It is because the adversarial attack may be more effective on the unknown classifiers than the known classifiers, typically when the known classifiers are more resistant to the attack.

%% file: sections/experiment.tex
We conducted a series of experiments to evaluated the transferability of multiple attacks against several forensic CNN architectures. Our database is created using 84,810 full-size JPEG images taken by 27 camera models from the Dresden Image Database~\cite{Dresden} (images are from 70 unique devices). We randomly selected 80\% for training, 10\% for validation and 10\% for testing. Next, we divided the full images into non-overlapping 256 by 256 image patches for each set. As a result, we ensure that there are no image patches from the same set coming from the same image and share the same statistics. To create the manipulated image patches, we selected three manipulations and five parameters that span a reasonable range for each manipulation. Then we manipulated each image patch in the database and obtained 15 unique sets of manipulated image patches. Along with the unaltered image classes, we obtained 16 classes corresponding to unaltered vs parameterized manipulations (manipulation parameterizer).  These images were also grouped into 4 classes of unaltered vs individual manipulations (manipulation classifier), and 2 classes of unaltered vs manipulated  (manipulation detector).  Table 1 shows the chosen manipulations and parameters we used to created manipulated image classes.  Due to computational cost constraints, we limited ourselves to three manipulations with five parameterizations each. Since we used 5 parameters per manipulation to create forged images, we in total created over 1,000,000 full sized JPEG images which are in bar with the up-to-date data size for training CNNs. 

\vspace{-0.5em}
\begin{table}[!h]
	\caption{Table 1: Editing operations and their associated parameters.}
	\label{tab: manipulations}
	\vspace{-0.5em}
\begin{center}
\begin{tabular}{|p{0.47\columnwidth}|p{0.45\columnwidth}|}
	\hline
	\textbf{Manipulations} & \textbf{Parameters}\\\hline
	Additive Gaussian white noise &$\mu=0,\sigma=0.5,1,1.5,2,2.5$\\\hline
	Gaussian blurring & $\sigma=1,1.5,2,2.5,3,3.5,4,4.5$\\\hline
	Median filtering &  window size$=3,5,7,9,11$\\\hline
\end{tabular}
\end{center}
\end{table}
\vspace{-3em}
\subsection{Baseline performance of forensic CNNs}
We started by training forensic CNNs using six CNN architectures and three class definitions. Each CNN was trained from scratch using stochastic gradient decent optimizer for 43 epochs and would stop early if validation accuracies started decreasing. The learning rate started with 0.0005 and would decrease to half every 4 epochs.  Batch size was 25. On average, we achieved 99.29\% accuracy using manipulation detector, 98.52\% for manipulation classifier, and 77.93\% for manipulation parameterizer.  These results are consistent with the state-of-art performance for manipulation detection. 
Table 2 demonstrates the classification accuracies achieved by trained manipulation detection CNNs. Each entry corresponds one pairing of CNN architecture and class definition. 

\vspace{-0.5em}
\begin{table}[!h]
	\caption{Table 2: Baseline classification accuracies achieved by six CNN architectures and three class definitions.}
	\vspace{-0.5em}
	\begin{center}
	\resizebox{0.48\textwidth}{!}{	
		\begin{tabular}{|c|c|c|c|}
			\hline
			\textbf{CNN Architect.} & \textbf{Manip. Detector} & \textbf{Manip. Classifier}  &\textbf{ Manip.  Parameterizer} 
			\\\hline
			\textit{MISLnet}& 99.84\% & 99.55\% & 86.24\% \\\hline
			\textit{TransferNet }& 99.20\%& 98.04\%& 65.27\% \\\hline
			\textit{PHNet}&99.58\% & 98.94\%&86.58\% \\\hline
			\textit{SRNet} & 99.16\%& 99.36\%  & 81.30\% \\\hline
			\textit{DenseNet}\textunderscore BC &98.13\%&95.66\%&65.50\%\\\hline
			\textit{VGG-19}& 99.87\% & 99.50\%&82.67\%\\\hline
		\textbf{Average}&  99.29\%&  98.51\%& 77.93\% \\\hline
	\end{tabular}}
	\end{center}
\end{table}
\vspace{-3em}
\subsection{Launching adversarial attacks}
We started by creating set of images used for evaluating the attacks.  From the testing set, we randomly selected 6,000 manipulated image patches that equally come from 15 manipulated image classes to form the \textit{ attack set.}  Then we used the two attack methods to attack each image patch in the attack set and targeted at ``unaltered" class. As a result, we obtained 216,000 anti-forensically attacked images.

For targeted I-FGSM,  we chose $\epsilon$ in equation to be $0.1$ and  the iteration to attack each image to be 100. For the GAN-based attack, we started by training a generator targeted at the ``unaltered'' class for each forensic CNNs, and then we used the trained generator to attack each image patch in the attack set. To train the generator, we randomly selected 360,000 manipulated image patches from the training set that equally come from 15 manipulated image. We trained the GAN-based attacked using the parameters in the original MISLGAN paper  authored by Chen et al~\cite{Mislgan}.  

\vspace{-0.8em}
\subsection{Baseline performance of adversarial attacks}
In this experiment, we would like to show the performance of the adversarial attacks against forensic CNNs when the attacks were trained directly to target at the ``unaltered" class of each forensic CNN. It corresponds to the scenario when the attacker has the perfect knowledge of investigator’s training data and full CNNs (i.e. including CNN architecture and the class definition).  

\vspace{-0.5em}
\begin{table}[!h]
	\caption{Table 3: Baseline performance of targeted  I-FGSM against forensic CNNs.}
	\vspace{-0.5em}
	\begin{center}
		\resizebox{0.48\textwidth}{!}{	
			\begin{tabular}{|c|c|c|c|}
				\hline
				\multicolumn{1}{|c|}{} & \multicolumn{3}{c|}{\textbf{Successful Attack Rate}} \\
				\cline{1-4}
				\textbf{CNN Architect.} & \textbf{Manip. Detector} & \textbf{Manip. Classifier}  &\textbf{ Manip.  Parameterizer} 
				\\\hline
				\textit{MISLnet}& 1.00& 1.00 & 1.00\\\hline
				\textit{TransferNet }& 0.99& 1.00& 1.00\\\hline
				\textit{PHNet}&0.87&0.96&1.00 \\\hline
				\textit{SRNet} & 0.88& 0.78& 1.00 \\\hline
				\textit{DenseNet}&0.63&0.98& 0.91\\\hline
				\textit{VGG-19}& 0.85&1.00&0.98\\\hline
				\textbf{Average} & 0.87 &  0.95 & 0.98 \\\hline
		\end{tabular}}
	\end{center}
	\caption{Table 4: Baseline performance of GAN-based attack against forensic CNNs.}
	\vspace{-0.5em}
	\begin{center}
		\resizebox{0.48\textwidth}{!}{	
			\begin{tabular}{|c|c|c|c|}
				\hline
				\multicolumn{1}{|c|}{} & \multicolumn{3}{c|}{\textbf{Successful Attack Rate} }\\
				\cline{1-4}
				\textbf{CNN Architect.} & \textbf{Manip. Detector} & \textbf{Manip. Classifier}  &\textbf{ Manip.  Parameterizer} \\\hline
				\textit{MISLnet}& 0.55 & 0.95& 0.84 \\\hline
				\textit{TransferNet }& 0.81 & 0.84& 0.98 \\\hline
				\textit{PHNet}&0.90 & 0.97&0.94\\\hline
				\textit{SRNet} & 0.88& 0.90 & 0.82\\\hline
				\textit{DenseNet}&0.90&0.94&0.94\\\hline
				\textit{VGG-19}& 0.71 &0.97&0.96\\\hline
				\textbf{Average} &  0.79& 0.93& 0.91 \\\hline
		\end{tabular}}
	\end{center}
\end{table}
\begin{table}[!h]
	\caption{Table 5: Transferability of targeted I-FGSM attack re-targeting on manipulation classifiers and parameterizers.}
		\vspace{-0.5em}
	\begin{center}
		\resizebox{0.48\textwidth}{!}{	
			\begin{tabular}{|c|c|c||c|c|}
				\hline
				\multicolumn{1}{|c|}{} & \multicolumn{2}{c||}{\textbf{ Successful Attack Rate} }& \multicolumn{2}{c|}{\textbf{Transferability Score}}\\
				\cline{1-5}
				\textbf{CNN Architect.} & \textbf{Manip. Classifier} & \textbf{Manip. Parameterizer}  &\textbf{Manip. Classifier} & \textbf{Manip. Parameterizer}  \\\hline
				\textit{MISLnet}& 0& 0 & 0 &0\\\hline
				\textit{TransferNet } & 0& 0 & 0 &0\\\hline
				\textit{PHNet}& 0& 0 & 0 &0\\\hline
				\textit{SRNet} & 0& 0 & 0 &0\\\hline
				\textit{DenseNet}& 0& 0 & 0 &0\\\hline
				\textit{VGG-19}& 0& 0 & 0 &0\\\hline
				\textbf{Average}& 0& 0& 0 & 0 \\\hline
		\end{tabular}}
	\end{center}
	
	\caption{Table 6: Transferability of targeted I-FGSM attack re-targeting on manipulation classifiers and parameterizers.}
		\vspace{-0.5em}
	\begin{center}
		\resizebox{0.48\textwidth}{!}{	
			\begin{tabular}{|c|c|c||c|c|}
				\hline
				\multicolumn{1}{|c|}{} & \multicolumn{2}{c||}{\textbf{ Successful Attack Rate} }& \multicolumn{2}{c|}{\textbf{Transferability Score}}\\
				\cline{1-5}
				\textbf{CNN Architect.} & \textbf{Manip. Detector} & \textbf{Manip. Parameterizer}  &\textbf{Manip. Detector} & \textbf{Manip. Parameterizer}  \\\hline
				\textit{MISLnet}& 0& 0 & 0 &0\\\hline
				\textit{TransferNet } & 0& 0 & 0 &0\\\hline
				\textit{PHNet}& 0& 0 & 0 &0\\\hline
				\textit{SRNet} & 0& 0 & 0 &0\\\hline
				\textit{DenseNet}& 0& 0 & 0 &0\\\hline
				\textit{VGG-19}& 0& 0 & 0 &0\\\hline
				\textbf{Average}& 0& 0& 0 & 0 \\\hline
		\end{tabular}}
	\end{center}
	\caption{Table 7: Transferability of targeted I-FGSM attack re-targeting on manipulation detectors and classifiers.}
		\vspace{-0.5em}
	\begin{center}
		\resizebox{0.48\textwidth}{!}{	
			\begin{tabular}{|c|c|c||c|c|}
				\hline
				\multicolumn{1}{|c|}{} & \multicolumn{2}{c||}{\textbf{ Successful Attack Rate} }& \multicolumn{2}{c|}{\textbf{Transferability Score}}\\
				\cline{1-5}
				\textbf{CNN Architect.} & \textbf{Manip. Detector} & \textbf{Manip. Classifier}  &\textbf{Manip. Detector} & \textbf{Manip. Classifier}  \\\hline
				\textit{MISLnet}& 0& 0 & 0 &0\\\hline
				\textit{TransferNet } & 0& 0 & 0 &0\\\hline
				\textit{PHNet}& 0& 0 & 0 &0\\\hline
				\textit{SRNet} & 0& 0 & 0 &0\\\hline
				\textit{DenseNet}& 0& 0 & 0 &0\\\hline
				\textit{VGG-19}& 0& 0 & 0 &0\\\hline
				\textbf{Average}& 0& 0& 0 & 0 \\\hline
		\end{tabular}}
	\end{center}

\caption{Table 8: Transferability of GAN-based attack re-targeting on manipulation classifiers and parameterizers.}
	\vspace{-0.5em}
\begin{center}
	\resizebox{0.48\textwidth}{!}{	
		\begin{tabular}{|c|c|c||c|c|}
			\hline
			\multicolumn{1}{|c|}{} & \multicolumn{2}{c||}{\textbf{ Successful Attack Rate} }& \multicolumn{2}{c|}{\textbf{Transferability Score}}\\
			\cline{1-5}
			\textbf{CNN Architect.} & \textbf{Manip. Classifier} & \textbf{Manip. Parameterizer}  &\textbf{Manip. Classifier} & \textbf{Manip. Parameterizer}  \\\hline
			\textit{MISLnet}& 0.004& 0.045 & 0.007 &0.082\\\hline
			\textit{TransferNet } & 0.008& 0.005 & 0.010 &0.006\\\hline
			\textit{PHNet}& 0.275& 0.120 & 0.306 &0.133\\\hline
			\textit{SRNet} & 0.420& 0.000 & 0.477 &0.000\\\hline
			\textit{DenseNet}& 0.005& 0.010& 0.008 &0.016\\\hline
			\textit{VGG-19}& 0.020& 0.090 & 0.024 &0.106\\\hline
			\textbf{Average}& 0.122 &0.045 & 0.139 & 0.057 \\\hline
	\end{tabular}}
\end{center}

	\caption{Table 9: Transferability of GAN-based attack re-targeting on manipulation detectors and parameterizers.}
		\vspace{-0.5em}
	\begin{center}
		\resizebox{0.48\textwidth}{!}{	
			\begin{tabular}{|c|c|c||c|c|}
				\hline
				\multicolumn{1}{|c|}{} & \multicolumn{2}{c||}{\textbf{ Successful Attack Rate} }& \multicolumn{2}{c|}{\textbf{Transferability Score}}\\
				\cline{1-5}
				\textbf{CNN Architect.} & \textbf{Manip. Detector} & \textbf{Manip. Parameterizer}  &\textbf{Manip. Detector} & \textbf{Manip. Parameterizer}  \\\hline
				\textit{MISLnet}& 0.090& 0.035 & 0.095 &0.037\\\hline
				\textit{TransferNet } & 0.000& 0.000 & 0.000 &0.000\\\hline
				\textit{PHNet}& 0.000 & 0.055 & 0.000 &0.057\\\hline
				\textit{SRNet} & 0.050& 0.005 & 0.056 &0.006\\\hline
				\textit{DenseNet}& 0.000& 0.000 & 0.000 &0.000\\\hline
				\textit{VGG-19} & 0.525& 0.260 & 0.541 &0.268\\\hline
				\textbf{Average}& 0.111& 0.059& 0.115& 0.060 \\\hline
		\end{tabular}}
	\end{center}
	
	\caption{Table 10: Transferability of GAN-based attack re-targeting on manipulation detectors and classifiers.}
		\vspace{-0.5em}
	\begin{center}
		\resizebox{0.48\textwidth}{!}{	
			\begin{tabular}{|c|c|c||c|c|}
				\hline
				\multicolumn{1}{|c|}{} & \multicolumn{2}{c||}{\textbf{ Successful Attack Rate} }& \multicolumn{2}{c|}{\textbf{Transferability Score}}\\
				\cline{1-5}
				\textbf{CNN Architecture} & \textbf{Manip. Detector} & \textbf{Manip. Classifier}  &\textbf{Manip. Detector} & \textbf{Manip. Classifier}  \\\hline
				\textit{MISLnet}& 0.365 & 0.035 & 0.435 &0.042\\\hline
				\textit{TransferNet } & 0.000 & 0.000 & 0.000 &0.000\\\hline
				\textit{PHNet}& 0.065& 0.490& 0.069 &0.521\\\hline
				\textit{SRNet} & 0.350& 0.440 & 0.427 &0.537\\\hline
				\textit{DenseNet}& 0.535& 0.135 & 0.588 &0.148\\\hline
				\textit{VGG-19}& 0.235& 0.185 & 0.240 &0.189\\\hline
				\textbf{Average}& 0.259& 0.214& 0.290 & 0.240 \\\hline
		\end{tabular}}
	\end{center}
\end{table}
Table 3 and Table 4 show the SARs we obtained for fooling forensic CNNs using I-FGSM and GAN-based attacks. Each entry is the individual SAR when targeting at a particular pair of CNN architecture and class definition.  One average, 
manipulation detectors can be fooled with 0.87 SAR using I-FGSM and 0.68 using GAN-based attack.
Manipulation classifiers can be fooled with 0.95 SAR using I-FGSM attack and 0.90 SAR using GAN-based attack. And manipulation parameterizers can be fooled with 0.98 SAR using I-FGSM and 0.91 SAR using GAN-based attack. First we noticed that under the perfect knowledge scenarios, both attacks can fool forensic CNNs with high SARs. Second, we noticed that for both attacks SARs for fooling manipulation detectors are consistently lower than the other two class definitions. For example,  targeted I-FGSM achieved 0.63 SAR on the manipulation detector using DenseNet architecture, compared to the over 0.90 SARs for fooling the other two class definitions. GAN-based attack achieved 0.55 SAR for fooling manipulation detector using MISLnet architecture, compared to over 0.85 SAR for fooling the other two class definitions. 
These results may imply that the manipulation detectors are more robust to adversarial attacks under the perfect knowledge scenario. 
\vspace{-1em}

\subsection {Transferability of adversarial attacks}
In this experiment, we evaluated the performance of the adversarial attacks against forensic CNNs when only the class definition of the target CNNs is changed. For each CNN architecture, we used forensic CNNs built with other class definitions to classify the adversarial images produced by individual attack. For example, if the adversarial images were produced to fool a MISLnet manipulation detector, we used the manipulation classifiers and parameterizers of MISLnet to classify these attacked images.

Table 5 - 10 show the successful attack rates and transferability scores achieved by the two adversarial attacks.  The left side of each table shows the SARs of fooling one particular pairing of CNN architecture and class definition, and the right side of each table shows the T-Scores of each class definition with respect to the trained class definition.   Table 5-7 shows that for targeted I-FGSM attack, both SARs and T-scores are 0's when re-targeting on different class definitions. It means the targeted I-FGSM attack cannot transfer to other class definitions. 

For GAN-based attack, the average SARs are less than 26\% and the average T-scores are less than 0.30. Shown in Table 8-10,  the GAN-based attacks can slightly transfer when trained with particular paring of class definitions and CNN architectures. Among the 36  transferability cases we tested, only 4 cases achieved over 0.5 T-scores and 20 cases are less then 0.1. The highest T-score was achieved when the GAN-based attack were trained to fool manipulation parameterizer using DenseNet architecture, then re-targeted at manipulation detectors.  However, there is still over 40\% SAR drop taken in account that class definition was the only changed factor.  These results demonstrated that adversarial attacks cannot transfer well across class definitions. Changing class definitions would significantly mitigate impact from adversarial attacks.

%% file: sections/conclusion.tex
In this paper, we investigated  the impact of class definitions on the transferability of adversarial attacks.While previous research has shown that the adversarial attacks cannot transfer across different CNN architectures or training data, we discovered that adversarial attacks are less transferable than previously thought. Particularly, by only changing the class definition of a forensic CNN,  we can significantly decrease the the performance of adversarial attacks. The finding holds consistent when using multiple adversarial attacks to attack many well-known CNN architectures. Besides, a secondary finding shows that some class definitions may be more robust to adversarial attacks than others. Particularly, the  SARs are lower when fooling binary detection under the perfect knowledge scenario.

%% file: sections/acknowledgement.tex
This material is based upon work supported by the National Science Foundation under Grant No. 1553610. Any opinions, findings, and conclusions or recommendations expressed in this material are those of the authors and do not necessarily reflect the views of the National Science Foundation.

This material is based on research sponsored by DARPA and Air Force Research Laboratory (AFRL) under agreement number PGSC-SC-111346-03. The U.S. Government is authorized to reproduce and distribute reprints for Governmental purposes notwithstanding any copyright notation thereon. The views and conclusions contained herein are those of the authors and should not be interpreted as necessarily representing the official policies or endorsements, either expressed or implied, of DARPA and Air Force Research Laboratory (AFRL) or the U.S. Government.